\documentclass[twoside,twocolumn]{article}

\usepackage{fitee}
\usepackage{subfigure}
\usepackage{soul}
\usepackage{color}
\usepackage{array}
\soulregister\cite7
\soulregister\citep7
\soulregister\ref7 

\usepackage[hidelinks, breaklinks = true]{hyperref}  % for DVI to PDF latex
\addto{\captionsenglish}{%
  
}

\begin{document}

%% article title
\title{A ground-based dataset and a diffusion model for on-orbit low-light image enhancement}

%% when all authors provide emails, no $\dagger$
\author[1]{Yi-man ZHU}%
\author[$\ddagger$1]{Lu WANG}
\author[1]{Jiny-yi YUAN}%
\author[1]{Yu GUO}%
\affil[1]{School of Automation, Nanjing University of Science and Technology,Xiaolingwei Street 200, Nanjing, 210094, China}

\shortauthor{Zhu et al.}	% one author: only Zhai; two authors: Zhai and Hu; three authors: Zhai et al.

\authmark{}

%% Only 1 affiliations
%\author{Wen-fei WANG}
%\author{Rong XIONG$^{\dagger\ddagger}$}
%\author{Jian CHU}
%\affil{\it\footnotesize State Key Laboratory of Industrial Control Technology, Institute of Cyber-Systems and Control,
%	\authorcr\it\footnotesize Zhejiang University, Hangzhou 310027, China}
          
% for authors, use \authorcr to begin with a new line, e.g., \author[2]{\authorcr Xian-liang HU}
% for the affiliation, use \authorcr\Affilfont\it to begin with a new line, e.g.,
%\affil[1]{Editorial Office of Journal of Zhejiang University SCIENCE,
%\authorcr\Affilfont\it Hangzhou 310027, Chin        

\corremailA{yiman@njust.edu.cn}
\corremailB{wanglu21@njust.edu.cn}
\corremailC{jingyi@njust.edu.cn}
\corremailD{guoyu@njust.edu.cn}
\emailmark{}	% when all authors provide emails, no \dagger

% abbrev. of month: Jan. Feb. Mar. Apr. May June July Aug. Sept. Oct. Nov. Dec.
\dateinfo{Received mmm.\ dd, 2016;	Revision accepted mmm.\ dd, 2016;    Crosschecked mmm.\ dd, 2017}

\abstract{On-orbit service is important for maintaining the sustainability of space environment. Space-based visible camera is an economical and lightweight sensor for situation awareness during on-orbit service. However, it can be easily affected by the low illumination environment. Recently, deep learning has achieved remarkable success in image enhancement of natural images, but seldom applied in space due to the data bottleneck. In this article, we first propose a dataset of the Beidou Navigation Satellite for on-orbit low-light image enhancement (LLIE). In the automatic data collection scheme, we focus on reducing domain gap and improving the diversity of the dataset. we collect hardware in-the-loop images based on a robotic simulation testbed imitating space lighting conditions. To evenly sample poses of different orientation and distance without collision, a collision-free working space and pose stratified sampling is proposed. Afterwards, a novel diffusion model is proposed. To enhance the image contrast without over-exposure and blurring details, we design a fused attention to highlight the structure and dark region. Finally, we compare our method with previous methods using our dataset, which indicates that our method has a better capacity in on-orbit LLIE}

% separate by semicolons

\doi{10.1631/FITEE.1000000}	% DOI of the paper, should be accurate
\code{A}
\clc{TP}

%\inpress	% uncomment this command if use "in press"

\publishyear{2018}
\vol{19}
\issue{1}
\pagestart{1}
\pageend{5}

%% when no funding, the following line should be removed, no period at last
\support{Project supported by the Postgraduate Research \& Practice Innovation Program (KYCX23\_0481) of Jiangsu Province, China.}

%\conf{A preliminary version of this paper has been presented at ??? Conference, date}
%\esm{Electronic supplementary materials: The online version of this article (http://dx/doi.org/10.1631/jzus.C1000000) contains supplementary materials, which are available to authorized users}
\orcid{Yi-man ZHU, http://orcid.org/0000-0002-7421-0188}	% corresponding author, or first author
\articleType{}
%\articleType can be `Science Letters:', `Review:', `Comment:', etc.
%Leave blank for research article.

\maketitle

\section{Introduction} \label{sec:introduction}
\begin{figure}[!t]
    \centering
	\subfigure[]{\includegraphics[width=0.32\columnwidth]{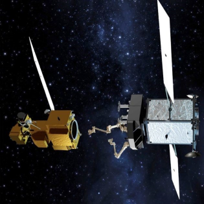}\label{zhu1_a}}
	\subfigure[]{\includegraphics[width=0.32\columnwidth]{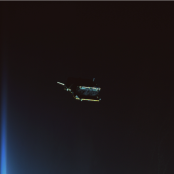}\label{zhu1_b}}
    \subfigure[]{\includegraphics[width=0.32\columnwidth]{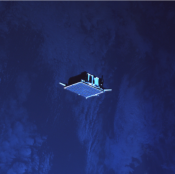}\label{zhu1_c}}
	\caption{The illustration of on-orbit service and low-loght images. (a) demonstrates service star refueling the target star. (b) and (c) are the real images of low-light and normal-light captured by a visible camera during PRISMA mission\citep{49}.}
    \label{zhu1}
\end{figure}

Due to the high frequency of space activities, the space environment has been seriously degraded by the by-product of these activities. As of February 2022, more than 25,000 space objects have been identified, including retired satellites, spacecraft, rocket bodies, and debris \citep{01}, which gravely threatens both the functioning and newly launched spacecraft \citep{02}. Hence, to ensure the long-term sustainability of space environment, clean up existing space debris and extend the lifespan of operational spacecraft, Active Debris Removal (ADR) and On-Orbit Servicing (OOS) have emerged as popular areas of research\citep{31}. Space robotic arm plays a vital role in performing various tasks associated with ADR and OOS, including capture and docking, repair and refurbishment, as demonstrated in Fig. \ref{zhu1_a}. To realize security and efficient robotic control, reliable sensors and data analysis are reqiured to provide better space situation awareness, especially when manipulating unkown targets\citep{03,32}. Space-based visible camera is one of the most essential sensors because of its following advantages: light in mass, compact in size, economical in power and informative in data \citep{04}. However, the variable illumination in space seriously affect the quality of the captured images, especially when the satellite is in the earth's shadow, as shown in Fig. \ref{zhu1_b}. The captured images can be extremely invisible, making it difficult to extract key features. In order to recover the buried details and improve data usability for downstream tasks and surveillance efficiency of SSA, we propose a novel framework to solve the problem of on-orbit low-light images enhancement (LLIE) 

Image enhancement methods based on classic theory \citep{15,16,36,35,34} have been widely used to improve the image contrast. Unfortunately, these methods are only suitable for situations where images already contain a good representation of the scene content. For extremely dark and noised images, they are likely to cause severe noise amplification after enhancement.

Deep learning methods have been applied to solve problems in various industries, demonstrating significant advancements\citep{14,47,48}. In LLIE, the mainstream methods can be classified as convolutional neural networks (CNNs) \citep{06,39,46,43,38,41} and generative adversarial networks (GANs) \citep{26,05,07,25,50}. CNN-based methods mainly adopt an encoder-decoder structure, but tend to directly replicate patterns and features rather than understanding the semantics. GANs can produce more accurate images trained with a generator and a discriminator competing with each other until reaching binary equilibrium. This mode is usually prone to collapse in the early stages of training.

Recently, diffusion models have brought the generative models to a new level \citep{08,24,51}. In contrast to GANs that lean on inner-loop maximization, diffusion models are guided on a concise and well-constructed loss, making it easier to convergence. Denoising diffusion probabilistic model (DDPM) represents a category of deep generative models that are based on (i) a forward diffusion stage, in which the input data is gradually perturbed over several steps by adding Gaussian noise, and (ii) a reverse denoising stage, in which a generative model is tasked at recovering the original input data from the diffused data by learning to gradually reverse the diffusion process \citep{09}. The generated images exihit very few artifacts and a high level of details. To date, diffuision models have been successfully applied to many image restoration tasks like image super-resolution\citep{19}, image inpainting \citep{20} and image enhancement \citep{37,40,42}.

However, most of these methods are based on public natural images. When brightening on-orbit low-light images, problems still remain , such as the insufficient enhancement, amplified noise and artificial features. As we all know, training data is imperative to the performance of deep learning models, which brings the demand for datasets in space research to the forefront. But, to the best of our knowledge, there is neither real-world nor synthetic dataset for on-orbit LLIE task because of the difficulty of capturing images in space. Besides, tasks in computer vision community pay more attention on the color fidelity and human-eye habits. For on-orbit mission, it is more important to save computing resources and provide accurate data for downstream tasks like visual measurement and estimation. Images of satellite in space also have different distribution because of its geometric shapes and metal surface. Therefore, it is more difficult to precisely reveal the structure and texture details from the dark region without generating more noises.

To solve these problems, in this article, we focus on a spacecraft with solar panel, taking the Beidou Navigation Satellite as an example. An automatic dataset collection scheme based on a 6-DoF robot is devised to tackle the data bottleneck for LLIE in space environment. Based on the dataset, we propose a novel attention-guided diffusion model to achieve the enhancement of on-orbit low-light satellite image, which is proved to be more effective than previous methods. The contributions of this paper have four aspects.
\begin{itemize}
    \item Unlike generating synthetic images as previous method, we first construct a dataset for on-orbit LLIE on a hardware in-the-loop testbed, which narrows the domain gap between real world and training set.
    \item To reduce manpower and ensure the safety and quality of data collection, we design a pose stratified sampling based on the collision-free working space obtained with physics engine.
    \item To our best knowledge, a novel diffusion model is proposed for low-light image enhancement in space environment for the first time. Compared with classic method and other state-of-the-art deep learning method, our method shows a better performance.
    \item Distinguished from existing diffusion model for image restoration, a fused-attention guidance is devised to extract information of illumination and structural distribution during the downsampling stage of training to guide light enhancement and detail preservation.

\end{itemize}

\section{Related Work}
\subsection{Space Dataset}
Images in space environment are scarce and hard to obtain, which hinders the application of intelligent method. Some efforts have been made to change this situation. The methods of acquiring datasets can be divided into two classes: capturing real images with different exposure and rendering images with physical engine. \cite{21} propose BUAA-SID1.0 database which uses 3ds Max to render images from full viewpoint based on the CAD model of 56 satellites. SPEED is the first publicly available machine learning dataset for spacecraft pose estimation, which consists of synthetic images generated with a non-physically-based render and a few real pictures taken in space mission\citep{10}. \cite{11} propose a realistic rendering dataset URSO, containing labeled images of spacecraft orbiting the earth. \cite{12} introduce SwissCube dataset created via physically-based rendering to reflect the illumination in space. \cite{22} provide a synthetic dataset SPARK, composed of 150k annotated multi-modal images aiming at space target recognition and detection. For generalized enhancement learning, \cite{46} introduce mapping functions to generate images of different exposure. Most datasets are synthetic because real images are extremely hard to obtain. The domain gap between synthetic images and real-scene images may lead to a relatively poor performance of the same architectures on real situations\citep{11,13}. Although some methods have used physically-based materials, the domain gap still exists. Besides, there is no public dataset for space object LLIE. In this article, we build a ground-test platform and an automatic collection scheme to make a hardware in-the-loop image dataset for LLIE, which practically addresses this problem.

\subsection{Low-light image enhancement}\label{subsec2}
The early works tackle LLIE can be typically categorized as the histogram equalization-based methods and the Retinex model-based methods. HE improves the contrast of an image by balancing the histogram of the entire image. Retinex model-based methods \citep{15,16,35,36,34} count on the assumption that a color image can be decomposed into reflectance and illumination. \cite{35} and \cite{34} improve the noise term in the Retinex model. \cite{16} transfer the illumination estimation into an optimization problem. With advancement of low-light data collection, deep learning methods have been proposed. Some researchers combine Retinex model with neural networks\citep{06,17}, with two CNNs modeling the illumination map and the reflectance map. But retinex-based methods depends on the decomposition assumption, which is not always valid and highly possible to lose fidelity and cause artifact\citep{14}. GANs have also been adopted in LLIE. \cite{05} extend the retinex network into GAN scheme. The image is firstly discomposed into two components then enhanced by two generator. \cite{07} propose EnlightenGAN, composed of a global-local discriminator structure and an attention-guided generator to handle the uneven light conditions. \cite{50} design an illumination-aware attention module to enhance the feature extraction. Based on the ability to capture global and local relationships of transformer framework, \cite{44} formulate a one-stage Retinex-based transformer to model the long-range dependences under the guidance of illumination information. \cite{45} utlize the YUV color space along with transformers to disentangling light and color information in images. Additionally, some researchers try to solve this problem based on different assumptions. \cite{43} construct WaveNet, which represents pixels as sample values of signal function and adaptively modulates the wave superposition mode with proposed Wave Transform Block. \cite{41} propose a trainable color space HVI, then design a two-branches network to decouple color and intensity. 

\subsection{Diffusion-based image restoration}
Recently, diffusion-based generative models have shown outstanding performance and widely used in image restoration tasks\citep{18}. \cite{19} propose SR3 and adapt DDPMs to conditional image generation, achieving a higher fidelity than GANs. \cite{20} found the potential of diffusion models in other tasks and propose a unified framework Palette for I2I translation tasks such as image inpainting and image colorization. \cite{40} propose a global structure-aware diffusion process with a curvature regularization term anchored in the intrinsic non-local structures of image data. \cite{42} propose a physics-based exposure model to start from a noisy image instead of pure noise in denoising process. \cite{37} uses a novel pyramid diffusion method to perform sampling in a pyramid resolution style to address the global degradation. \cite{52} propose FusionDiff, fusing several locally focused source images to obtain globally clear images with DDPM. \cite{53} propose a real-time underwater image enhancement (UIE) by applying a novel sampling inference strategy on DDPM. \cite{54} design a DDPM with color compensation as a conditional guide to address UIE. \cite{55} train an unconditional diffusion model prior on the joint space of color and depth to restore underwater image. Although many diffuision-based models have emerged, there is no effective solution for on-orbit low-light image enhancement.

\section{Data Acquisition}
Image datasets in space are difficult to collect especially those for LLIE training. Because it is impossible to capture the real picture of different illuminance with the same state of targets and surroundings when the sensor is mounted on the moving chaser. In this article, a 6-DoF robot will carry the target model simulating spin motion and a camera will collect data of different exposure. To realize automatic collection while avoiding collision and ensuring unbiasedness, we first construct a simulation environment in pybullet 3.2.5 to select suitable poses for the robot. Then the pair-image data will be collected on the ground-test platform. To achieve a data collection procedure without human intervention, the robot arm should decide which pose to perform and plan a trajectory. A collision-free working space and pose stratified sampling method are proposed to ensure diversity and unbiasedness, as well as avoiding robot collision with satellite model and itself.

\subsection{The ground-test platform}

\begin{figure}[h]%
\centering
\includegraphics[width=\columnwidth]{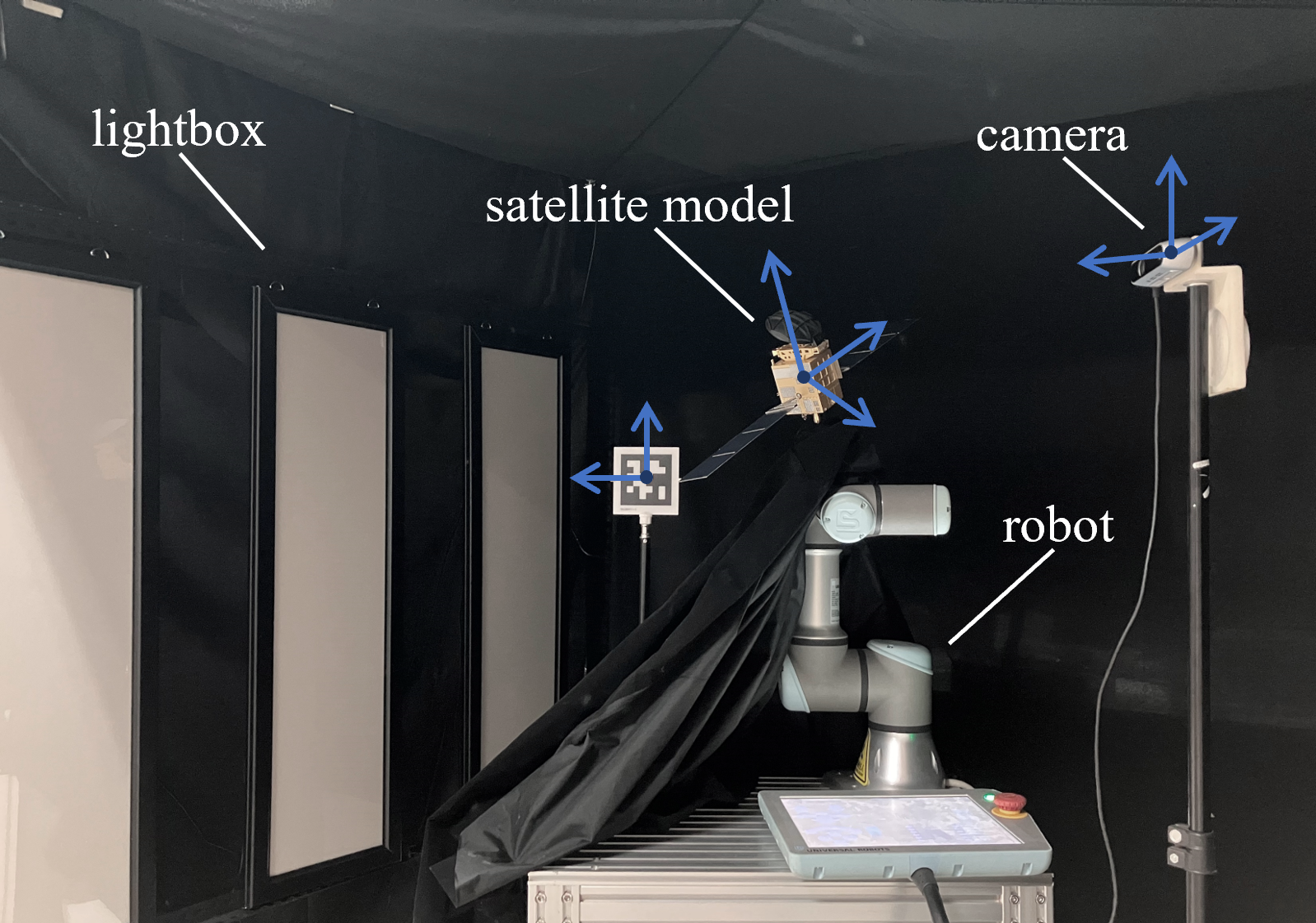}
\caption{The layout of the ground-test platform. t is composed of a 6 DoF robot carrying a satellite model spinning and a calibrated camera.}\label{fig2}
\end{figure}

The platform consists of a 6 DoF Universal Robot UR3 and a reduced-scale simulate metal models of Beidou Navigation Satellite. The robot arm carries the model to simulate random motion in space. The metal surface of the model generate strong reflection as the real one in space. To avoid collision and singular posture if possible, the motion is set to simple spinning during collection, which will not affect the data diversity. To create a space luminous environment of high-fidelity, the room is surrounded by black absorbing materials. 3 led light-boxes are placed to provide the Earth's diffuse light of different angles and intensity. The overall settings are shown in Fig.~\ref{fig2}. 

\subsection{Collision-free working space}
A robot working space defines the space range of task execution. The primary and secondary working space of robot have been proposed to describe the space which can be reached from one direction and multiple directions\citep{33}. However, these working spaces have not consider the following motion after reaching a target point. In this paper, we propose a collision-free working space, which enables the robot arm move from initial pose to target pose and simulate satellite spinning without collision. The procedure is as follows.

\begin{figure}[!t]
	\centering
	\subfigure[The robot model]{\includegraphics[width=0.49\columnwidth]{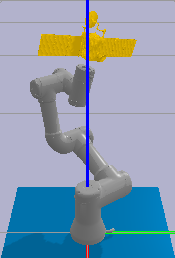}\label{zhu2_a}}
	\subfigure[The collision mesh]{\includegraphics[width=0.49\columnwidth]{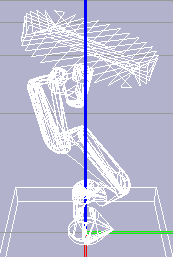}\label{zhu2_b}}
	\caption{The virtual environment in Pybullet. On the left is the visual rendered model, on the right is the collision mesh of the model.}
	\label{zhu2}
  \end{figure}

\noindent
\textbf{1) Set collision geometry.} We apply the Bullet physic engine for python and modify the UR3 model file. The satellite model is added as the last link with a fixed joint. The collision geometry of satellite is specifically set to cylindrical to avoid collision during spinning. The virtual environment is shown in Fig.~\ref{zhu2_a}. and Fig.~\ref{zhu2_b}

\noindent
\textbf{2) Generate candidate poses.} \cite{23} calculated the robot space with Monte Carlo method, which tries to understand a system by generating a large number of random samples. However, it can be time-consuming to reach convergence. To speed up the sampling, we apply the Quasi Monte Carlo method based on Halton sequence to generate candidate poses.

\noindent
\textbf{3) Check collision and arrival.} According to the candidate pose, a joint-space trajectory planning is adopted and during the execution, we check whether the robot collides with the table and the satellite through the Bullet collection detection library.

\noindent
\textbf{4) Finish the loop.} If no collision occurs and the robot reaches target poses with a small error tolerance, the poses are added to the collision-free working space. The trajectory is recorded and the robot moves along the trajectory to the initial pose.

\begin{figure}[!t]
	\centerline{\includegraphics[width=\columnwidth]{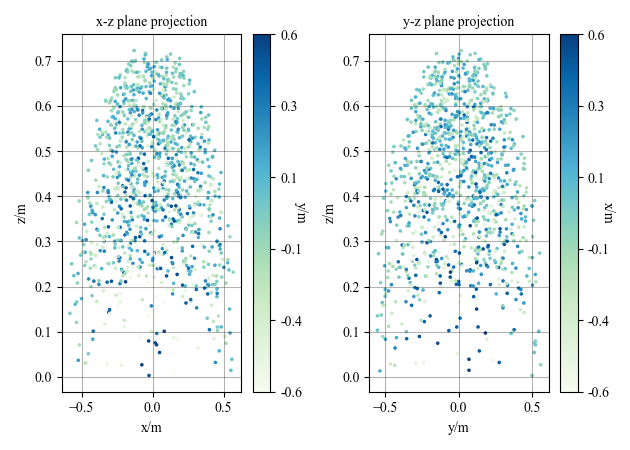}}
	\caption{The $x-z$ and $y-z$ sectional view of the collision-free working space.}
	\label{zhu3}
\end{figure}

The space constructed is shown in Fig.~\ref{zhu3}., any points in this space can promise security autonomous data collection.


\subsection{Pose stratified sampling}
After constructing the collision-free working space, we select poses from the space. In an ideal situation, the more poses are sampled, the more uniform and extensive the dataset are. However, considering the limited computing resource, we tend to improve the representativeness of samples as much as possible with a limit size of dataset. With random sampling, the sampled poses may cluster in a certain part because it requires more sample times to approximate the entire distribution. Therefore, we adopt stratified sampling, which converges faster than random sampling. 

\begin{figure}[!t]
  \centering
  \centerline{\includegraphics[width=\columnwidth]{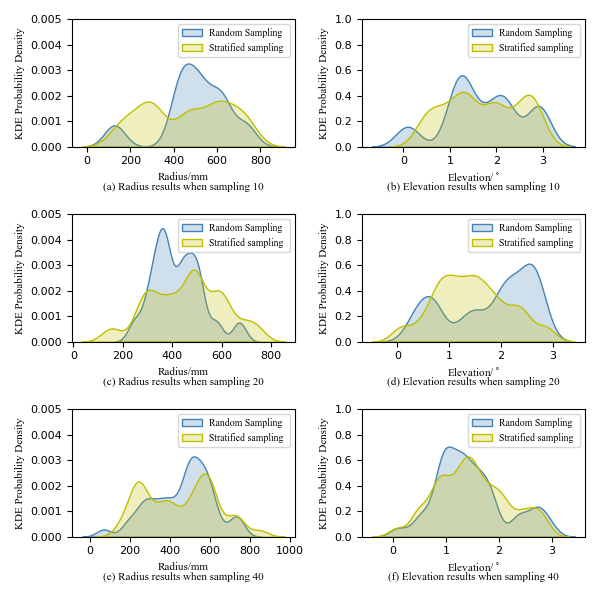}}
  \caption{The comparison of radius and elevation of different sampling times. The first column is the radius of the camera in the satellite spherical coordinates after different times of sampling. The second column is the elevation of the camera in the satellite spherical coordinates after different times of sampling.}
  \label{zhu4}
\end{figure}

First, the working space is divided by radius, azimuth angle and elevation angle in spherical coordinates. Then, the poses are sent to the robot controller via socket. Suppose the initial angle of the terminal joint is denoted by $q_0$. The robot carries the satellite move around the z-axis of the terminal joint, with a camera capturing photos at $10^\circ$ intervals. Then we calculate the spherical coordinates of the calibrated camera optical center relative to the satellite. The comparison of random sampling and stratified sampling with 10, 20 and 40 times of sampling are shown in Figs.~\ref{zhu4}. As illustrated, when the sampling time is small, the random sampling results cluster in a certain interval. But the distribution of stratified sampling results is robust to the sampling times. Finally, we select 20 times of pose sampling and get 720 pairs of images of the satellite.

\subsection{Data preprocessing}
\begin{figure}[!t]
    \centerline{\includegraphics[width=\columnwidth]{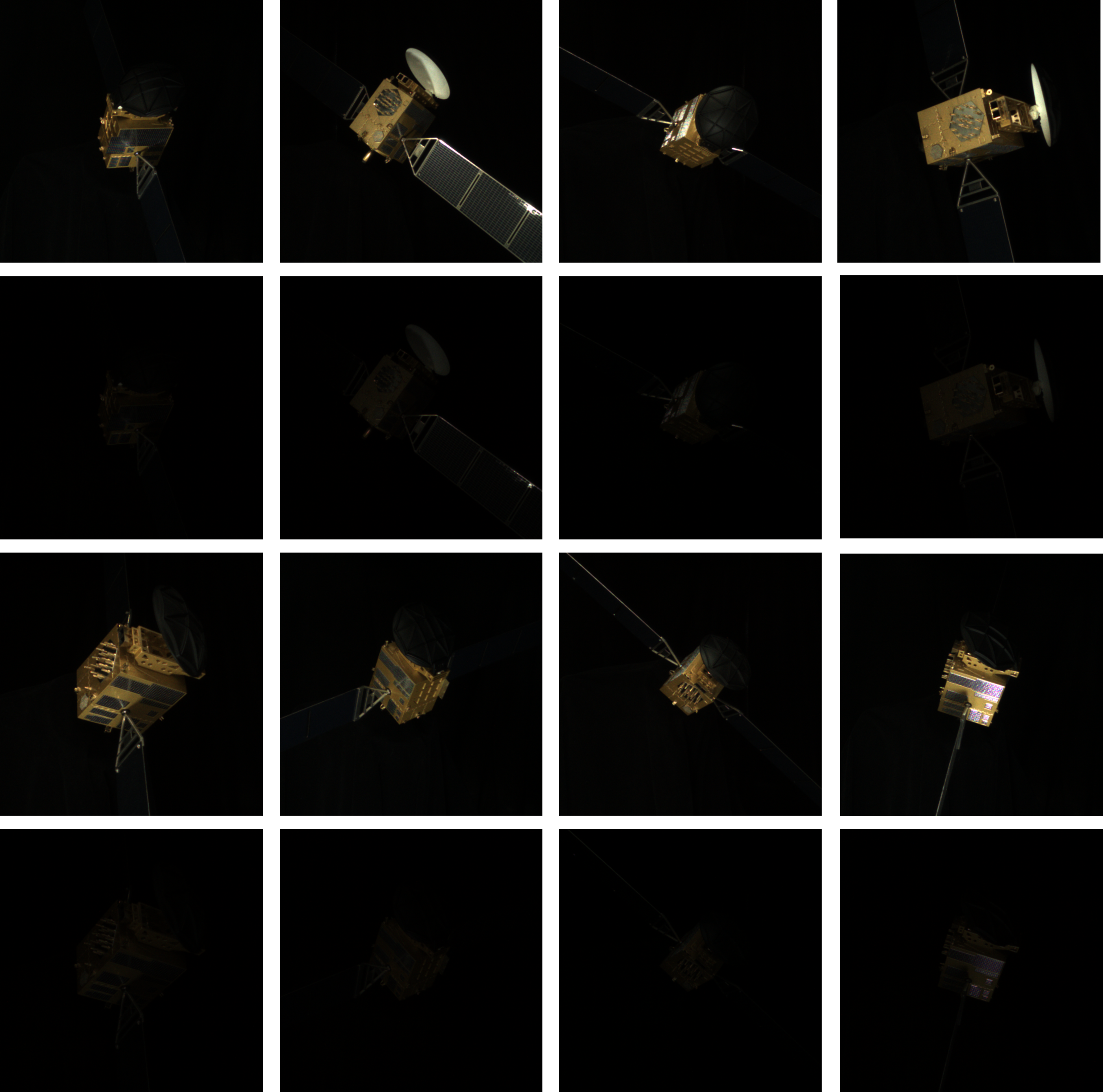}}
    \caption{Images in our dataset. In the first and third rows are the normal-light images. In the second and fourth rows are the low-light images.}
    \label{zhu6}
\end{figure}

 The camera exposure time is set as 156 $\mu$s and 1248 $\mu$s respectively to capture high and low illumination image of the same scene. The pictures are captured with a Realsense D435 sensor as a format of Bayer array to avoid data compression, especially under low illumination. The Bayer arrays are converted to RGB images with Bilinear Interpolation. Designate a value in Bayer array as $I_{i,j}$ and the corresponding pixel value in RGB image is $T_{i,j} = \left( R,G,B\right)$. When  $I_{i,j}$ represents red channel, $R$ equals $I_{i,j}$. $G$ and $B$ are calculated from the four green and blue channels in the neighboring areas of $I_{i,j}$. The calculation formula is as follows
\begin{equation}
    \begin{aligned}
        R&=I_{i,j},
        \\
        G&=\frac{I_{i-1,j}+I_{i+1,j}+I_{i,j-1}+I_{i,j+1}}{4},
        \\
        B&=\frac{I_{i-1,j-1}+I_{i+1,j+1}+I_{i-1,j+1}+I_{i+1,j-1}}{4}.
    \end{aligned} 
\end{equation}
When $I_{i,j}$ represents blue channel, $B$ equals $I_{i,j}$. $R$ and $G$ are calculated from the four red and green channels in the neighboring areas of $I_{i,j}$. The calculation formula is as follows
\begin{equation}
    \begin{aligned}
    R&=\frac{I_{i-1,j-1}+I_{i+1,j-1}+I_{i-1,j+1}+I_{i+1,j+1}}{4},
    \\
    G&=\frac{I_{i,j-1}+I_{i,j+1}+I_{i-1,j}+I_{i+1,j}}{4},
    \\
    B&=I_{i,j}.
    \end{aligned} 
\end{equation}
When $I_{i,j}$ represents green channel and $I_{i,j-1}$ represents red, there are only two red and blue channels in the $I_{i,j}$ neighbourhood. The calculation formula is as follows
\begin{equation}
    \begin{aligned}
        R=\frac{I_{i,j-1}+I_{i,j+1}}{2}, G=I_{i,j}, B=\frac{I_{i-1,j}+I_{i+1,j}}{2}.
    \end{aligned} 
\end{equation}
When $I_{i,j}$ represents green channel and $I_{i,j-1}$ represents blue, The calculation formula is same as before
\begin{equation}
    \begin{aligned}
        R=\frac{I_{i-1,j}+I_{i+1,j}}{2}, G=I_{i,j}, B=\frac{I_{i,j-1}+I_{i,j+1}}{2}.
    \end{aligned} 
\end{equation}
After converting the Bayer Array to RGB formats, a center cropping is adopted to reshape the image to $ 640 \times 640$. Samples of the dataset are displayed in Fig.~\ref{zhu6}. 

\section{LLIE for Satellite Image}
\subsection{The diffusion model}
\begin{figure}[!t]
	\centerline{\includegraphics[width=\columnwidth]{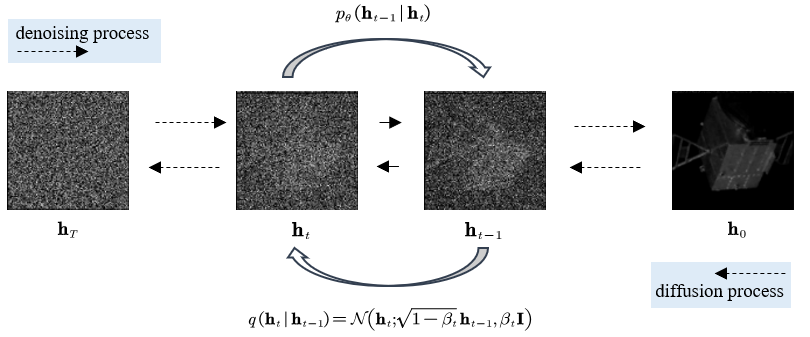}}
	\caption{The process of diffusion model. In the forward diffusion process, random Gaussian noise $\epsilon$ controlled by the timestep is gradually added to the ground truth image to until completely turning the image into noise $\mathbf{h_T}$. At the same time, a neural network is trained to predict the Gaussian noise. In the reverse process, $\epsilon$ at each step is estimated to recover $\mathbf{h_T}$ from the noise. }
	\label{zhu7}
  \end{figure}

Diffusion models comprise a forward diffusion process and a reverse denoising process, as illustrated in Fig.~\ref{zhu7}. The formulation of diffusion model based LLIE problem is demonstrated. Define a pair of low-light image and high light image $\left( \mathbf{l},\mathbf{h} \right) $. Noise is added to $\mathbf{h}$ for $T$ times until turning it into an isotropic two-dimensional Gaussian noise. The noised $\mathbf{h}$ at $t$ step is designated as $ \mathbf{h}_t \left( \mathbf{h}_0=\mathbf{h} \right) $, and the noise $\epsilon $ at $t$ step is scheduled by hyper-parameters $\beta _t$, $\left( 0<\beta _1<\beta _2<...<\beta _T\ll 1 \right) $ . The Markovian chain is described as follows:
\begin{equation}
    \begin{aligned}
        q\left( \mathbf{h}_t\mid \mathbf{h}_{t-1} \right) &=\mathcal{N} \left( \mathbf{h}_t;\sqrt{1-\beta _t}\mathbf{h}_{t-1},\beta _t\mathbf{I} \right) 
        \\
        q\left( \mathbf{h}_{1:T}\mid \mathbf{h}_0 \right) &=\Pi _{t=1}^{T}q\left( \mathbf{h}_t\mid \mathbf{h}_{t-1} \right). 
    \end{aligned} 
    \label{eq6}
\end{equation}
Based on the above property, a closed form of $\mathbf{h}_t$, for any intermediate $t$ is obtained 
\begin{equation}
    \begin{aligned}
        q\left( \mathbf{h}_t\mid \mathbf{h}_0 \right) =\mathcal{N} \left( \mathbf{h}_t;\sqrt{\gamma _t}\mathbf{h}_0,\left( 1-\gamma _t \right) \mathbf{I} \right),
    \end{aligned} 
    \label{eq7}
\end{equation}
where $\alpha _t=1-\beta _t$ and $\gamma _t=\prod{_{s=1}^{t}\alpha _s}$.

Since $\beta _t$ is much smaller than 1, the reverse process can also be considered as a Markovian chain. In order to recover $\hat{\mathbf{h}}_0$ from noise, the posterior probability $p\left(\mathbf{h}_{t-1} \mid \mathbf{h}_{t} \right) $ must be achieved, which is difficult to calculate. So we attempt to use network to estimate it.
\begin{equation}
    \begin{aligned}
        p_{\theta}\left( \mathbf{h}_{t-1}\mid \mathbf{h}_t \right) =\mathcal{N} \left( \mathbf{h}_{t-1};\mu _{\theta}\left( \mathbf{l},\mathbf{h}_t,t \right) ,\Sigma _{\theta}\left( \mathbf{l},\mathbf{h}_t,t \right) \right) .
    \end{aligned} 
    \label{eq8}
\end{equation}
Following DDPM \citep{08}, a similar variational distribution $q\left( \mathbf{h}_{1:T}\mid \mathbf{h}_{0} \right)$ is introduced. By deriving logarithmic likelihood function and variation analysis, the problem of calculating $p\left( \mathbf{h}_{t-1} \mid \mathbf{h}_{t}\right)$  is transformed into approximating the posterior probability distribution with $q\left( \mathbf{h}_{t-1}\mid \mathbf{h}_t,\mathbf{h}_0 \right)$, which is tractable.

\begin{equation}
    \begin{aligned}
        q\left( \mathbf{h}_{t-1}\mid \mathbf{h}_t,\mathbf{h}_0 \right) =\mathcal{N} \left( \mathbf{h}_{t-1};\tilde{\mu}_t\left( \mathbf{l},\mathbf{h}_t,\mathbf{h}_0 \right) ,\tilde{\beta}_t\mathbf{I} \right),
    \end{aligned} 
    \label{eq9}
\end{equation}
where
\begin{equation}
    \begin{aligned}
        \tilde{\mu}_t&=\frac{\sqrt{\alpha _t}\left( 1-\gamma _{t-1} \right)}{1-\gamma _t}\mathbf{h}_t+\frac{\sqrt{\gamma _{t-1}}\left( 1-\alpha _t \right)}{1-\gamma _t}\mathbf{h}_0,
        \\
        \tilde{\beta}_t&=\frac{\left( 1-\gamma _{t-1} \right) \left( 1-\alpha _t \right)}{1-\gamma _t}.
    \end{aligned} 
    \label{eq10}
\end{equation}
$\tilde{\beta}_t$ is already known. $\mathbf{h}_0$  has a relationship with $\mathbf{h}_t$ as described in Eq.\ref{eq7}, so $\tilde{\mu}_t$ is formulated as 
\begin{equation}
    \begin{aligned}
        \tilde{\mu}_t&=\frac{1}{\sqrt{\alpha _t}}\left( \mathbf{h}_t-\frac{1-\alpha _t}{\sqrt{1-\gamma _t}}\epsilon \right).
    \end{aligned} 
    \label{eq11}
\end{equation}
Use network $f_{\theta}$ to estimate the noise, and achieve
\begin{equation}
    \begin{aligned}
        \mu _{\theta}\left( \mathbf{l},\tilde{\mathbf{h}}_t,\gamma _t \right) =\frac{1}{\sqrt{\alpha _t}}\left( \mathbf{h}_t-\frac{1-\alpha _t}{\sqrt{1-\gamma _t}}f_{\theta}\left( \mathbf{l},\tilde{\mathbf{h}}_t,\gamma _t \right) \right),
    \end{aligned} 
    \label{eq12}
\end{equation}
where $f_{\theta}\left( \mathbf{l},\tilde{\mathbf{h}}_t,\gamma _t \right)$ conditions on input low-light image $\mathbf{l}$, noisy image $\tilde{\mathbf{h}}$, and current noise level $\gamma _t$, The ground truth of noise is recorded in diffusion process. Therefore, The network is trained to predict noise $\epsilon$ by optimizing

\begin{equation}
    \begin{aligned}
        \mathbb{E} _{(l,h)}\mathbb{E} _{\epsilon ,\gamma}\left\| f_{\theta}(l,\tilde{\mathbf{h}},\gamma )-\epsilon \right\| _{p}^{p}.
    \end{aligned} 
    \label{eq13}
\end{equation}

  \begin{figure*}[!t]
	\centerline{\includegraphics[width=163mm]{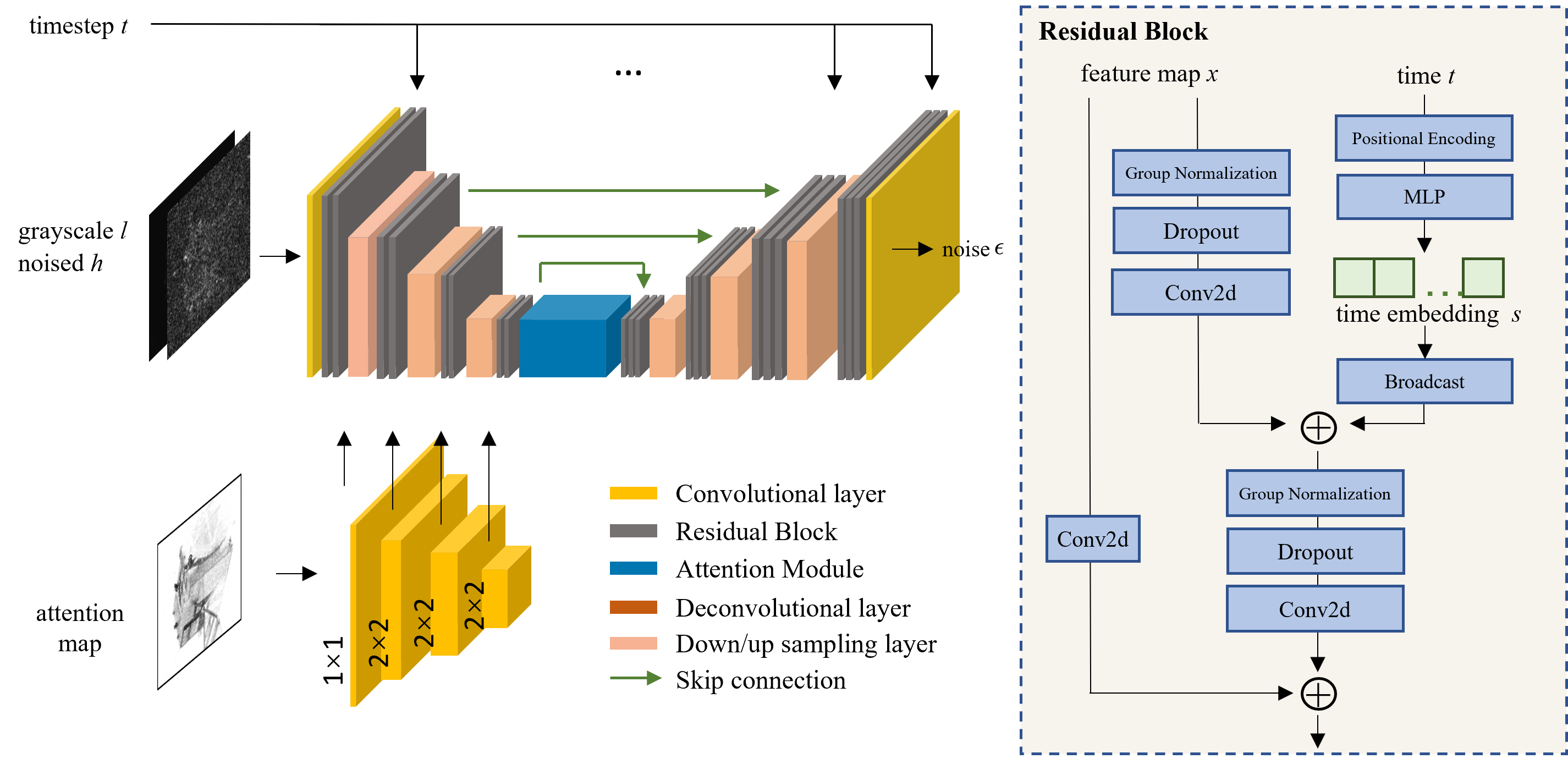}}
	\caption{The network structure. The network takes the grayscale $\mathbf{l}$, $\mathbf{h_t}$ and $t$ as input. $\mathbf{l}$ is the low-light image to be enlightened. $\mathbf{h_t}$ is the noised normal-light image at $t$ step in forward process. The network follows a U-Net structure. The timestep $t$ is fed into a sinusoidal position encoding and multi-layer perceptrons to achieve a time embedding $s$. Then $s$ is boardcast and fed into each residual block. The FAG of $\mathbf{l}$ are reshaped by convolution layers to different scale and inserted to each down sampling layer. The network outputs the noise $\epsilon$ supervised by the noise sampled in the forward process.}
	\label{zhu8}
  \end{figure*}

  \begin{figure}[!t]
	\centerline{\includegraphics[width=\columnwidth]{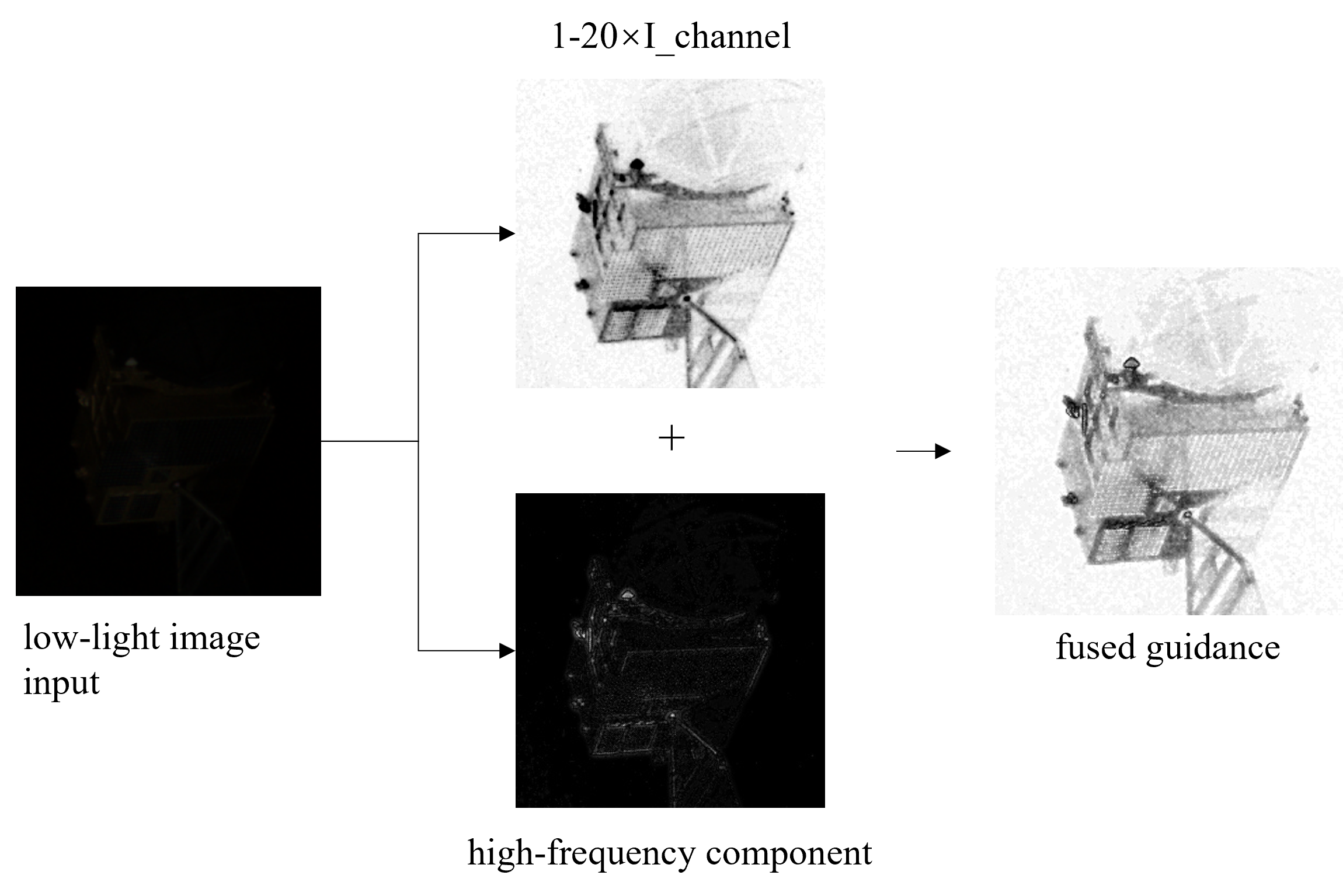}}
	\caption{Fusing the intensity channel with high-frequency component to obtain guidance map}
	\label{att}
  \end{figure}

\subsection{Fused attention guidance}

In computer vision community, RGB or Raw input are mostly considerred. However, the uneven prediction of RGB channel results in patterns' color shift \citep{20,37}. To avoid error accumulation and save calculation, we directly predict the grayscale image of the enhanced results. 
Images of satellite in space have a severely-uneven illuminance distribution because of the geometric shapes and special surface materials. To improve the enhancement, we propose a fused attention guidance (FAG) to highlight the detailed information such as edges and textures of images and guide the network focus on relighting the dark region of images. Suppose an input low-light image $\mathbf{l}(u,v)$, its guidance map is calculated as
\begin{equation}
    \begin{aligned}
		\textnormal{FAG} &= [ 1- \frac{\lambda }{\sqrt{3} } (R+G+B) ]\\
		 &+ \mathcal{FFT}^{-1}\{ H(u,v)\cdot \mathcal{FFT}(u,v)   \}
    \end{aligned} 
    \label{eq13}
\end{equation}
 The first part is the reverse of the intensity channel after converting low-light images into HSI color space, which guides the network to enhance the dark aeras. $\lambda $ is a parameters controlled by the degree of image darkness. When the image is extemely dark, the intensity channel is very weak, $\lambda $ should be used to highlight the information in intensity channel. The second part corresponds to the high-frequency component of low-light images, obtained by a high-pass filter after fast Fourier transform. As shown in Fig.~\ref{att}, the value is empirically set to be 20, the textures areas are strengthened while the high reflective areas are suppressed.

\subsection{Network structure}
The network is comprised of an encoder and a decoder with residual blocks as the core building block. The skip connections between the encoder and decoder are designed to ensure the reuse rate of features with the same dimension. We find that using pooling layers for downsampling and deconvolution layers instead of the nearest neighbor interpolation for upsampling achieves a better result. To insert the FAG into the pipeline, we rescale the map with 4 convolutional layers and fuse the feature map into the corresponding downsampling layers, as shown in Fig.~\ref{zhu8}. In addition to retaining feature information, the residual block embeds the time information $t$. The timestep $t$ is put through sinusoidal position encoding and multi-layer perceptrons to achieve a time embedding $s$. Then $s$ is boardcast to add with the image feature maps and fed into each residual block to make the network conditional on $t$, which is a significant factor for predicting the noise of different diffusion stage.

\section{Experiment}
\begin{figure*}[!t]
	\centerline{\includegraphics[width=16.4cm]{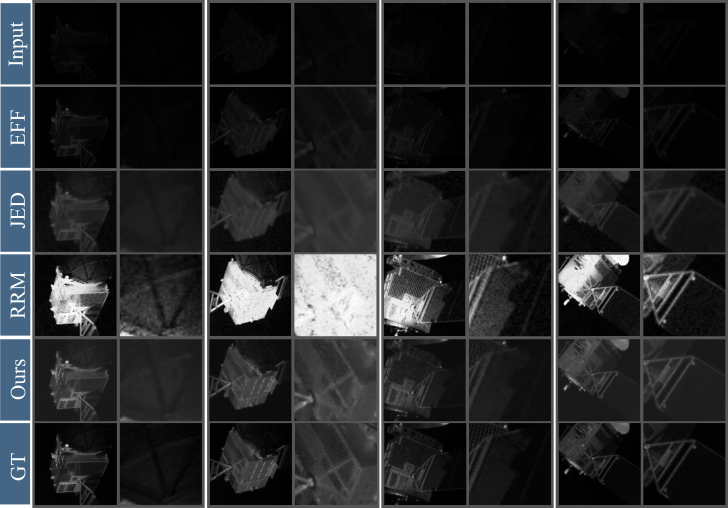}}
	\caption{Visual comparison of the enhanced results and enlarged local details with different traditional methods EFF \citep{36} JED \citep{34}, RRM \citep{35}. }
	\label{zhu10}
\end{figure*}

\begin{figure*}[!t]
	\centerline{\includegraphics[width=16.4cm]{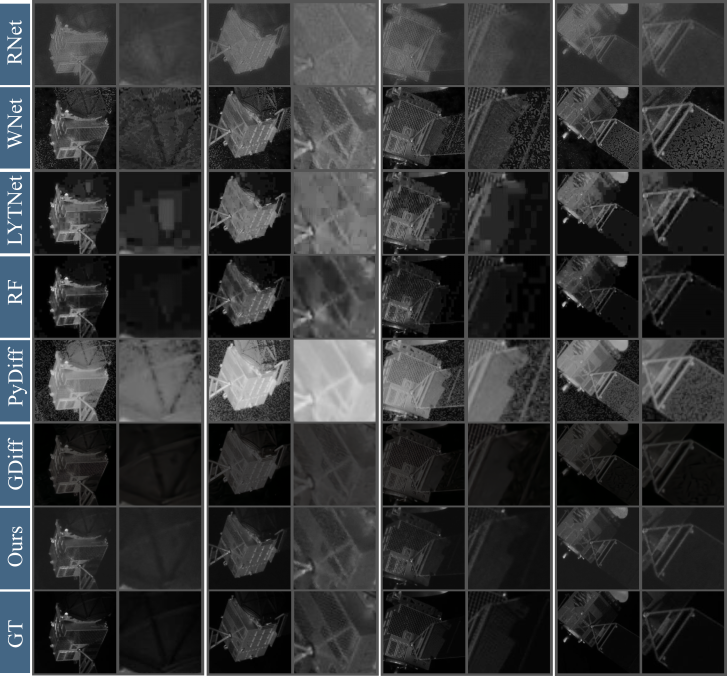}}
	\caption{Visual comparison of the enhanced results and enlarged local details with different deep learning methods, including CNN-based RetinexNet \citep{06} and WaveNet \citep{43}, transformer-based RetinexFormer \citep{44} and LYTNet \citep{45}, diffusion-based PyramidsDiff \citep{37} and GlobalDiff \citep{40}.}
	\label{zhu10_}
\end{figure*}

\subsection{Implementation details}
\begin{figure}[h]%
    \centering
    \includegraphics[width=\columnwidth]{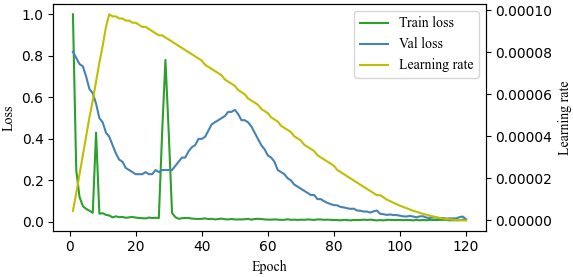}
    \caption{The curves of the learning rate, training loss and validation loss.}\label{loss}
\end{figure}

The entire framework is programmed with Pytorch \citep{29} 1.12.0. Experiments are run on an NVIDIA TITAN RTX GPU with an Intel Xeon E5-2678 CPU. The dataset has a total of 720 pairs of images, with 700 pairs for training and 20 pairs for testing. The input images are rescaled and cropped to $256\times256$ and augmented with flipping and rotation. During our training, the total timestep $T$ is $2\times10^{3} $. The noise weight $\beta _t$ is arranged by cosine schedule with an offset of $8\times10^{-3}$ to provide a linear drop-off of weight in the middle of the process, while changing very little near both ends to prevent abrupt changes in noise level \citep{24}. The network is trained for 100 epoches and aimed at reducing a simple $l_2$ loss. To avoid early over-fitting, we apply a dynamic learning rate $lr_i$ changing over iteration step $i$ scheduled by a linear warm-up \citep{27} and decays following cosine annealing \citep{28} after reaching $lr=1\times10^{-4}$. The dynamics of the training process are demonstrated in Fig. \ref{loss}. In the early stages of training, loss oscillations and overfitting exist, as the learning rate decreases and dropout takes effect, the network keeps learning steadily and gradually converges.

\subsection{On-orbit LLIE effect analysis}

\begin{table*}[thp]\footnotesize
    \addtolength{\tabcolsep}{-6pt}
	\centering
	\caption{Comparison with previous method} \label{Table1}
	\addtolength{\tabcolsep}{4.8pt}
	\begin{tabular*}{16.05cm}{l|ccc|cccccc|c|}
		\toprule[0.75pt]
		\multirow{3}{*}{Method} & \multicolumn{3}{|c}{Traditional method} & \multicolumn{7}{|c|}{Deep learning method}  \\
		\cmidrule[0.5pt]{2-4}\cmidrule[0.5pt]{5-11} & EFF & JED & RRM & RNet & PyDiff & WNet & GDiff & RFormer & LYTNet & Ours              \\
        & ICCVW'17 & ISCAS'18 & TIP'18  & BMVC'18 & IJCAI'23 & PG'23 & ICCV'23 & NIPS'23 & Arxiv'24 &\\
		\midrule[0.5pt]
		PSNR$\uparrow$& 19.21 & \underline{24.25} & 9.284 & 15.05 & 10.51 & 18.03 & 17.47 & 23.83 & 18.31 &\textbf{25.14}            \\
		SSIM$\uparrow$& 0.5499 & 0.5958 & 0.3697 & 0.3368 & 0.2601 & 0.4413 & 0.3487 & \textbf{0.7602} & 0.5503 &\underline{0.6107}             \\
		FSIM$\uparrow$& 0.8263 & 0.8043 & 0.6231 & 0.8292 & 0.6689 & 0.7677 & 0.6610 & \underline{0.8899} & 0.8144 &\textbf{0.9102}               \\
		LPIPS$\downarrow$& 0.2159 & 0.3072 & 0.5062 & 0.3278 & 0.5192 & 0.3578 & 0.4987 &\underline{0.2043} & 0.3496& \textbf{0.0951}            \\
		\bottomrule[0.75pt]
		\multicolumn{11}{p{15.6cm}}{\scriptsize $^*$ The optimal values are in bold and the suboptimal values are underlined}
	\end{tabular*}
\end{table*}
To demonstrate that our method has better ability in enlighten on-orbit images, we compare the results with different methods, including traditional methods EFF \citep{36} JED \citep{34}, RRM \citep{35}, deep learning methods RetinexNet \citep{06} and WaveNet \citep{43}, RetinexFormer \citep{44} and LYTNet \citep{45}, PyramidsDiff \citep{37} and GlobalDiff \citep{40}. Methods are evaluated with image quality assessment PSNR,SSIM,FSIM and LPIPS. The comparison results are illustrated in Table~\ref{Table1}. Our method is better than previous results and competitive with SOTA. Fig.~\ref{zhu10} demonstrate the comparison with classic methods. The EFF and RRM have problems of insufficient and over enhancement. JED enlarges the noise in dark region and blurs details while increasing the illumination. Fig.~\ref{zhu10_} illustrates the effect of deep learning methods. RetinexNet (Rnet) and WaveNet (WNet) cause varing degrees detail blurring and image distortion. LYTNet and RetinexFormer (RF) have a good capcity of learning the distribution of target domain, but the image resolution is degraded. Compared with other newly-proposed diffusion-based method, our method generates less noises and artifacts while brightening the low-light images. 

\subsection{The ablation study}
\begin{figure}[!t]
	\centerline{\includegraphics[width=\columnwidth]{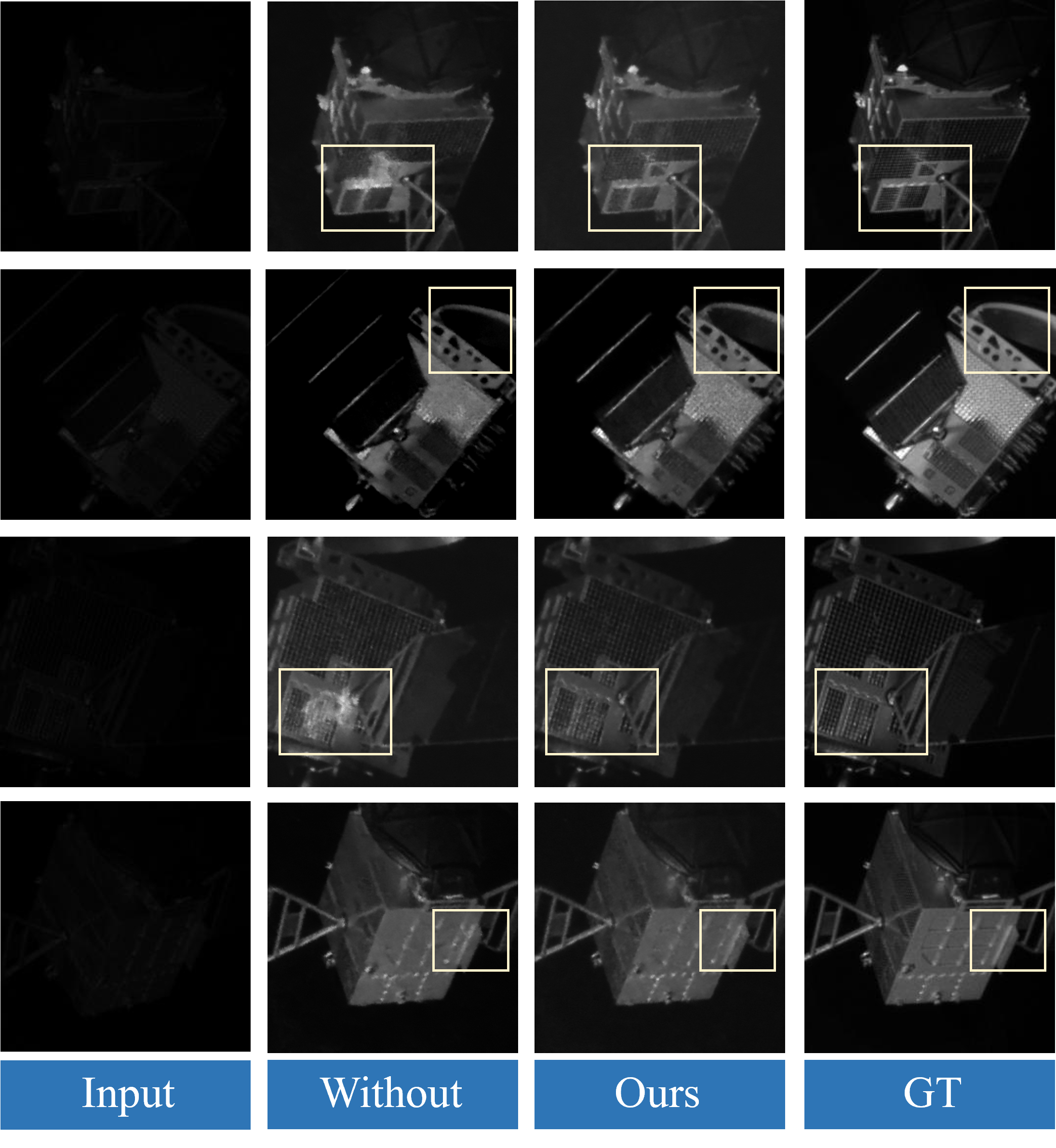}}
	\caption{Visual comparison of the enhanced results with and without FAG guidance. The yellow boxes highlight aeras with significant improvement.}
	\label{zhu9}
  \end{figure}

We carry out an ablation study to verify the effectiveness of the FAG. The enlightened images with and without FAG are compared in Fig.~\ref{zhu9}. The first column are the input low-light images. The second column are results without FAG, which have unreal artifacts on texture and fail to enlighten extreme dark areas. The results in the third column are generated with FAG. The evaluation metrics of ablation study are displayed in Table~\ref{table2}, in which we can conclude that the quality is improved significantly with FAG on feature extraction during down sampling. The visual results and metrics indicate that our method has a better capacity in low-light enhancement and fidelity maintenance. 

\begin{figure}[!t]
	\centerline{\includegraphics[width=\columnwidth]{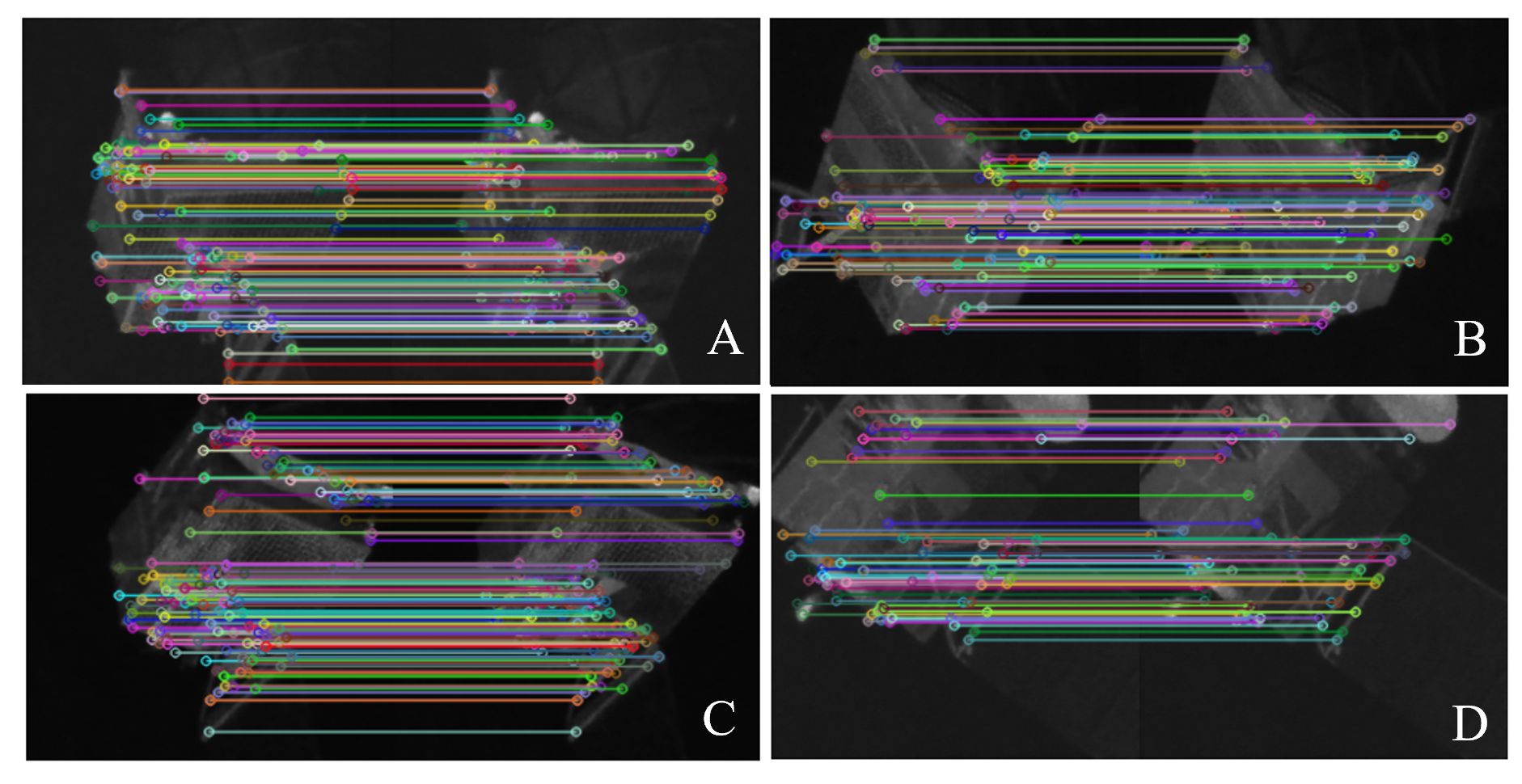}}
	\caption{The sift matching result of enhanced images of two continuous frames}
	\label{sift}
\end{figure}

\begin{table}[t]
	\centering
	\caption{The ablation study}
	\label{table2}
	  \begin{tabular}{lcc}
		\toprule[0.75pt]
		Method		& without guidance & with guidance \\
		\midrule[0.5pt]
		PSNR$\uparrow$ & 23.81 & \textbf{25.14}   \\
		SSIM$\uparrow$ & 0.5937 & \textbf{0.6107}  \\
		FSIM$\uparrow$ & 0.8994 &  \textbf{0.9102}  \\
		LPIPS$\downarrow$ & 0.1206 & \textbf{0.0951}    \\
		\bottomrule[0.75pt]
	\end{tabular}
\end{table}
\begin{table}[b]
	\centering
	\caption{The number of good SIFT feature}
	\label{Table3}
	  \begin{tabular}{lccccc}
		\toprule[0.75pt]
				& A & B & C & D & Average\\
		\midrule[0.5pt] 
		Raw & 1 & 0 & 0 & 0  & 0.25\\
		Enhanced & 140 & 130 & 159 & 83 & 128\\
		\bottomrule[0.75pt]
	\end{tabular}
\end{table}

\subsection{Engineering applicability analysis}
After a contrast enhancement, the low-light images can provide more information in following high level visual tasks, which is vital for the success of on-orbit tasks. To prove that the features of the enhanced images become richer and preserve consistency, we performed SIFT feature point matching on the enhanced images of continuous frames, as shown in Fig.~\ref{sift}. Table~\ref{Table3} shows these SIFT matching results of enhanced images compared with raw images. In low-light images, features can barely be extracted, after our enhancement, more features are extracted.

\section{Conclusion}
In this article, the on-orbit LLIE problem of a Beidou Navigation Satellite in extreme dark light conditions is explored. To provide dataset of high quality for deep learning methods, we propose an automatic dataset collection scheme to build a pair dataset of satellite image for deep learning. The scheme considers the diversity of the datasets and reduces the domain gap between dataset and real world, which lays the foundation for the data-driven methods. Based on the dataset, a novel diffusion model is proposed. To enhance the image contrast without over-exposure and blurring details, we design FAG which is an attention map highlighting the structure and dark region. The experimental results indicate that our approach shows better performance than SOTA on-orbit LLIE.  However, the reverse process of diffusion models can take a long time during inference, how to reduce the reverse steps and ensure the quality of generation at the same time is of great importance. In order to migrate this method to multiple satellite targets, few-shot learning technique can be explored in the future. Besides, the space environment is variant, except for low-light image enhancement, other image restoration problems such as motion blur, over-exposure caused by direct sunlight are hoped to be resolved by multi-task diffusion model.
\\
\\
{\noindent{\textbf{Contributors~{\tiny{(refer to https://www.casrai.org/credit.html)}}}}

{\footnotesize Yi-man ZHU designed the method. Yi-man ZHU and Jing-yi YUAN  carried out the comparation experiments. Yi-man ZHU drafted the manuscript. Lu Wang and Yu GUO helped organize the manuscript. Yi-man ZHU and Lu WANG revised and finalized the paper.
}
~

{\noindent{\textbf{Compliance with ethics guidelines}}

{\footnotesize Yi-man ZHU, Lu WANG, Jing-yi YUAN and Yu GUO declare that they have no conflict of interest.
}

%\balance
\bibliographystyle{fitee}
\bibliography{reference}

\begin{thebibliography}{}\footnotesize
\itemsep -4pt
\vspace{-8pt}

\bibitem[\protect\astroncite{Brateanu {et~al.}}{2024}]{45}
Brateanu A, Balmez R, Avram A, {et~al.}, 2024.
\newblock Lyt-net: Lightweight yuv transformer-based network for low-light image enhancement.
\newblock {\em arXiv preprint arXiv:240115204}.

\bibitem[\protect\astroncite{Cai {et~al.}}{2023}]{44}
Cai Y, Bian H, Lin J, {et~al.}, 2023.
\newblock Retinexformer: One-stage retinex-based transformer for low-light image enhancement.
\newblock Proceedings of the IEEE/CVF International Conference on Computer Vision (ICCV), p.12504-12513.

\bibitem[\protect\astroncite{Chen {et~al.}}{2024}]{51}
Chen X, Liu Z, Xie S, {et~al.}, 2024.
\newblock Deconstructing denoising diffusion models for self-supervised learning.
\newblock {\em arXiv preprint arXiv:240114404}. \\ https://doi.org/10.48550/arXiv.2401.14404

\bibitem[\protect\astroncite{Chen {et~al.}}{2018}]{26}
Chen YS, Wang YC, Kao MH, {et~al.}, 2018.
\newblock Deep photo enhancer: Unpaired learning for image enhancement from photographs with gans.
\newblock Proceedings of the IEEE Conference on Computer Vision and Pattern Recognition, p.6306-6314. \\ https://doi.org/10.1109/CVPR.2018.00660

\bibitem[\protect\astroncite{Civardi {et~al.}}{2023}]{32}
Civardi GL, Bechini M, Quirino M, {et~al.}, 2023.
\newblock Generation of fused visible and thermal-infrared images for uncooperative spacecraft proximity navigation.
\newblock {\em Advances in Space Research}. \\ https://doi.org/10.1016/j.asr.2023.03.022

\bibitem[\protect\astroncite{Cowardin and Miller}{2022}]{01}
Cowardin H, Miller R, 2022.
\newblock The intentional destruction of cosmos 1408.
\newblock {\em Orbital Debris Quarterly News}.

\bibitem[\protect\astroncite{Croitoru {et~al.}}{2023}]{09}
Croitoru FA, Hondru V, Ionescu RT, {et~al.}, 2023.
\newblock Diffusion models in vision: A survey.
\newblock {\em IEEE T PATTERN ANAL}. \\ https://doi.org/10.1109/TPAMI.2023.3261988

\bibitem[\protect\astroncite{Dang {et~al.}}{2023}]{43}
Dang J, Li Z, Zhong Y, {et~al.}, 2023.
\newblock Wavenet: Wave-aware image enhancement.
\newblock Pacific Graphics Short Papers and Posters, p.1-9. \\ https://doi.org/10.2312/pg.20231267

\bibitem[\protect\astroncite{Diao {et~al.}}{2011}]{04}
Diao HF, Li Z, Ma Zh, 2011.
\newblock Simulation of space-based visible surveillance images for space surveillance.
\newblock International Symposium on Photoelectronic Detection and Imaging 2011: Space Exploration Technologies and Applications,  8196:94-102. \\ https://doi.org/10.1117/12.899120

\bibitem[\protect\astroncite{Feng {et~al.}}{2024}]{41}
Feng Y, Zhang C, Wang P, {et~al.}, 2024.
\newblock You only need one color space: An efficient network for low-light image enhancement.
\newblock {\em arXiv preprint arXiv:240205809}. \\ https://doi.org/10.48550/arXiv.2402.05809

\bibitem[\protect\astroncite{Fu {et~al.}}{2022}]{50}
Fu Y, Hong Y, Chen L, {et~al.}, 2022.
\newblock Le-gan: Unsupervised low-light image enhancement network using attention module and identity invariant loss.
\newblock {\em Knowledge-Based Systems}. \\ https://doi.org/10.1016/j.knosys.2021.108010

\bibitem[\protect\astroncite{Guan {et~al.}}{2024}]{54}
Guan M, Xu H, Jiang G, {et~al.}, 2024.
\newblock Diffwater: Underwater image enhancement based on conditional denoising diffusion probabilistic model.
\newblock {\em IEEE Journal of Selected Topics in Applied Earth Observations and Remote Sensing}. \\ https://doi.org/10.1109/JSTARS.2023.3344453

\bibitem[\protect\astroncite{Guo {et~al.}}{2016}]{16}
Guo X, Li Y, Ling H, 2016.
\newblock Lime: Low-light image enhancement via illumination map estimation.
\newblock {\em IEEE T IMAGE PROCESS}. \\ https://doi.org/10.1109/TIP.2016.2639450

\bibitem[\protect\astroncite{Gupta}{1986}]{33}
Gupta KC, 1986.
\newblock On the nature of robot workspace.
\newblock {\em The International Journal of Robotics Research}. \\ https://doi.org/10.1177/027836498600500212

\bibitem[\protect\astroncite{Harris {et~al.}}{2021}]{03}
Harris C, Thomas D, Kadan J, {et~al.}, 2021.
\newblock Expanding the space surveillance network with spacebased sensors using metaheuristic optimization techniques.
\newblock Advanced Maui Optical and Space Surveillance Technologies Conference (AMOS), Maui, HI, USA, p.1-13.

\bibitem[\protect\astroncite{Ho {et~al.}}{2020}]{08}
Ho J, Jain A, Abbeel P, 2020.
\newblock Denoising diffusion probabilistic models.
\newblock {\em Advances in Neural Information Processing Systems}. \\ https://doi.org/10.48550/arXiv.2006.11239

\bibitem[\protect\astroncite{Hou {et~al.}}{2023}]{40}
Hou J, Zhu Z, Hou J, {et~al.}, 2023.
\newblock Global structure-aware diffusion process for low-light image enhancement.
\newblock Advances in Neural Information Processing Systems,  36:79734-79747.

\bibitem[\protect\astroncite{Hu {et~al.}}{2021}]{12}
Hu Y, Speierer S, Jakob W, {et~al.}, 2021.
\newblock Wide-depth-range 6d object pose estimation in space.
\newblock Proceedings of the IEEE/CVF Conference on Computer Vision and Pattern Recognition, p.15870-15879. \\ https://doi.org/10.48550/arXiv.2104.00337

\bibitem[\protect\astroncite{Huang {et~al.}}{2023}]{38}
Huang Z, Li J, Zhen H, {et~al.}, 2023.
\newblock Filter-cluster attention based recursive network for low-light enhancement.
\newblock {\em Front Inform Technol Electron Eng}. \\ https://doi.org/10.1631/FITEE.2200344

\bibitem[\protect\astroncite{Jiang {et~al.}}{2021}]{07}
Jiang Y, Gong X, Liu D, {et~al.}, 2021.
\newblock Enlightengan: Deep light enhancement without paired supervision.
\newblock {\em IEEE T IMAGE PROCESS}. \\ https://doi.org/10.1109/TIP.2021.3051462

\bibitem[\protect\astroncite{Kisantal {et~al.}}{2020}]{10}
Kisantal M, Sharma S, Park TH, {et~al.}, 2020.
\newblock Satellite pose estimation challenge: Dataset, competition design, and results.
\newblock {\em IEEE T AERO ELEC SYS}. \\ https://doi.org/10.1109/TAES.2020.2989063

\bibitem[\protect\astroncite{Ledkov and Aslanov}{2022}]{02}
Ledkov A, Aslanov V, 2022.
\newblock Review of contact and contactless active space debris removal approaches.
\newblock {\em PROG AEROSP SCI}. \\ https://doi.org/10.1016/j.paerosci.2022.100858

\bibitem[\protect\astroncite{Li {et~al.}}{2021}]{14}
Li C, Guo C, Han L, {et~al.}, 2021.
\newblock Low-light image and video enhancement using deep learning: A survey.
\newblock {\em IEEE T PATTERN ANAL}. \\ https://doi.org/10.1109/TPAMI.2021.3126387

\bibitem[\protect\astroncite{Li {et~al.}}{2016}]{23}
Li J, Zhao F, Li X, {et~al.}, 2016.
\newblock Analysis of robotic workspace based on monte carlo method and the posture matrix.
\newblock 2016 IEEE International Conference on Control and Robotics Engineering (ICCRE), p.1-5. \\ https://doi.org/10.1109/ICCRE.2016.7476145

\bibitem[\protect\astroncite{Li {et~al.}}{2018}]{35}
Li M, Liu J, Yang W, {et~al.}, 2018.
\newblock Structure-revealing low-light image enhancement via robust retinex model.
\newblock {\em IEEE Trans Image Process}. \\ https://doi.org/10.1109/TIP.2018.2810539

\bibitem[\protect\astroncite{Li {et~al.}}{2024}]{52}
Li M, Pei R, Zheng T, {et~al.}, 2024.
\newblock Fusiondiff: Multi-focus image fusion using denoising diffusion probabilistic models.
\newblock {\em Expert Systems With Applications}. \\ https://doi.org/10.1016/j.eswa.2023.121664

\bibitem[\protect\astroncite{Li and Zhang}{2019}]{39}
Li N, Zhang J, 2019.
\newblock Automatic image enhancement by learning adaptive patch selection.
\newblock {\em Front Inform Technol Electron Eng}. \\ https://doi.org/10.1631/FITEE.1700125

\bibitem[\protect\astroncite{Loshchilov and Hutter}{2016}]{28}
Loshchilov I, Hutter F, 2016.
\newblock Sgdr: Stochastic gradient descent with warm restarts.
\newblock {\em arXiv preprint arXiv:160803983}. \\ https://doi.org/10.48550/arXiv.1608.03983

\bibitem[\protect\astroncite{Lu {et~al.}}{2023}]{53}
Lu S, Guan F, Zhang H, {et~al.}, 2023.
\newblock Speed-up ddpm for real-time underwater image enhancement.
\newblock {\em IEEE Transactions on Circuits and Systems for Video Technology}. \\ https://doi.org/10.1109/TCSVT.2023.3314767

\bibitem[\protect\astroncite{Ma and Yarats}{2021}]{27}
Ma J, Yarats D, 2021.
\newblock On the adequacy of untuned warmup for adaptive optimization.
\newblock Proceedings of the AAAI Conference on Artificial Intelligence,  35(10):8828-8836. \\ https://doi.org/10.48550/arXiv.1910.04209

\bibitem[\protect\astroncite{Mithun {et~al.}}{2023}]{31}
Mithun P, Manu HN, Mini CR, {et~al.}, 2023.
\newblock Active debris removal: A review and case study on leopard phase 0-a mission.
\newblock {\em ADV SPACE RES}. \\ https://doi.org/10.1016/j.asr.2023.06.015

\bibitem[\protect\astroncite{Musallam {et~al.}}{2021}]{22}
Musallam MA, Gaudilliere V, Ghorbel E, {et~al.}, 2021.
\newblock Spacecraft recognition leveraging knowledge of space environment: simulator, dataset, competition design and analysis.
\newblock 2021 IEEE International Conference on Image Processing Challenges (ICIPC), p.11-15. \\ https://doi.org/10.1109/ICIPC53495.2021.9620184

\bibitem[\protect\astroncite{Nathan {et~al.}}{2024}]{55}
Nathan Opher N, Levy D, Treibitz T, {et~al.}, 2024.
\newblock Osmosis: Rgbd diffusion prior for underwater image restoration.
\newblock {\em arXiv preprint arXiv:240314837}. \\ https://doi.org/doi.org/10.48550/arXiv.2403.14837

\bibitem[\protect\astroncite{Nichol and Dhariwal}{2021}]{24}
Nichol AQ, Dhariwal P, 2021.
\newblock Improved denoising diffusion probabilistic models.
\newblock International Conference on Machine Learning, p.8162-8171. \\ https://doi.org/10.48550/arXiv.2102.09672

\bibitem[\protect\astroncite{Park {et~al.}}{2023}]{13}
Park TH, M{\"a}rtens M, Jawaid M, {et~al.}, 2023.
\newblock Satellite pose estimation competition 2021: Results and analyses.
\newblock {\em ACTA ASTRONAUT}. \\ https://doi.org/10.1016/j.actaastro.2023.01.002

\bibitem[\protect\astroncite{Paszke {et~al.}}{2019}]{29}
Paszke A, Gross S, Massa F, {et~al.}, 2019.
\newblock Pytorch: An imperative style, high-performance deep learning library.
\newblock {\em Advances in neural information processing systems}. \\ https://doi.org/10.48550/arXiv.1912.01703

\bibitem[\protect\astroncite{Proen{\c{c}}a and Gao}{2020}]{11}
Proen{\c{c}}a PF, Gao Y, 2020.
\newblock Deep learning for spacecraft pose estimation from photorealistic rendering.
\newblock 2020 IEEE International Conference on Robotics and Automation (ICRA), p.6007-6013. \\ https://doi.org/10.1109/ICRA40945.2020.9197244

\bibitem[\protect\astroncite{Rahman {et~al.}}{1996}]{15}
Rahman Zu, Jobson DJ, Woodell GA, 1996.
\newblock Multi-scale retinex for color image enhancement.
\newblock Proceedings of 3rd IEEE international conference on image processing,  3:1003-1006. \\ https://doi.org/10.1109/ICIP.1996.560995

\bibitem[\protect\astroncite{Rao {et~al.}}{2021}]{05}
Rao N, Lu T, Zhou Q, {et~al.}, 2021.
\newblock Seeing in the dark by component-gan.
\newblock {\em IEEE SIGNAL PROC LET}. \\ https://doi.org/10.1109/LSP.2021.3079848

\bibitem[\protect\astroncite{Ren {et~al.}}{2020}]{34}
Ren X, Li M, Cheng W, {et~al.}, 2020.
\newblock Joint enhancement and denoising method via sequential decomposition.
\newblock 2018 IEEE International Symposium on Circuits and Systems (ISCAS), p.1-5. \\ https://doi.org/10.1109/ISCAS.2018.8351427

\bibitem[\protect\astroncite{Saharia {et~al.}}{2022a}]{20}
Saharia C, Chan W, Chang H, {et~al.}, 2022a.
\newblock Palette: Image-to-image diffusion models.
\newblock ACM SIGGRAPH 2022 Conference Proceedings, p.1-10. \\ https://doi.org/10.1145/3528233.3530757

\bibitem[\protect\astroncite{Saharia {et~al.}}{2022b}]{19}
Saharia C, Ho J, Chan W, {et~al.}, 2022b.
\newblock Image super-resolution via iterative refinement.
\newblock {\em IEEE T PATTERN ANAL}. \\ https://doi.org/10.1109/TPAMI.2022.3204461

\bibitem[\protect\astroncite{Triantafyllidou {et~al.}}{2020}]{25}
Triantafyllidou D, Moran S, McDonagh S, {et~al.}, 2020.
\newblock Low light video enhancement using synthetic data produced with an intermediate domain mapping.
\newblock Computer Vision--ECCV 2020: 16th European Conference, Glasgow, UK, August 23--28, 2020, Proceedings, Part XIII 16, p.103-119. \\ https://doi.org/10.48550/arXiv.2007.09187

\bibitem[\protect\astroncite{Wang {et~al.}}{2019}]{17}
Wang Y, Cao Y, Zha ZJ, {et~al.}, 2019.
\newblock Progressive retinex: Mutually reinforced illumination-noise perception network for low-light image enhancement.
\newblock Proceedings of the 27th ACM international conference on multimedia, p.2015-2023. \\ https://doi.org/10.1145/3343031.3350983

\bibitem[\protect\astroncite{Wang {et~al.}}{2023}]{42}
Wang Y, Yu Y, Yang W, {et~al.}, 2023.
\newblock Exposurediffusion: Learning to expose for low-light image enhancement.
\newblock 2023 IEEE/CVF International Conference on Computer Vision (ICCV), p.12404-12414. \\ https://doi.org/10.1109/ICCV51070.2023.01143

\bibitem[\protect\astroncite{Wei {et~al.}}{2018}]{06}
Wei C, Wang W, Yang W, {et~al.}, 2018.
\newblock Deep retinex decomposition for low-light enhancement.
\newblock {\em arXiv preprint arXiv:180804560}. \\ https://doi.org/10.48550/arXiv.1808.04560

\bibitem[\protect\astroncite{Xu {et~al.}}{2019}]{49}
Xu Y, Feng K, Yan X, {et~al.}, 2019.
\newblock Robust features extraction for on-board monocular-based spacecraft pose acquisition.
\newblock AIAA Scitech 2019 Forum, p.1-15. \\ https://doi.org/10.2514/6.2019-2005

\bibitem[\protect\astroncite{Xu {et~al.}}{2021}]{46}
Xu Y, Yang C, Sun B, {et~al.}, 2021.
\newblock A novel multi-scale fusion framework for detail-preserving low-light image enhancement.
\newblock {\em Information Sciences}. \\ https://doi.org/10.1016/j.ins.2020.09.066

\bibitem[\protect\astroncite{Xu {et~al.}}{2023a}]{48}
Xu Y, Feng K, Yan X, {et~al.}, 2023a.
\newblock Cfcnn: A novel convolutional fusion framework for collaborative fault identification of rotating machinery.
\newblock {\em Information Fusion}. \\ https://doi.org/10.1016/j.inffus.2023.02.012

\bibitem[\protect\astroncite{Xu {et~al.}}{2023b}]{47}
Xu Y, Yan X, Sun B, {et~al.}, 2023b.
\newblock Online knowledge distillation based multiscale threshold denoising networks for fault diagnosis of transmission systems.
\newblock {\em IEEE Transactions on Transportation Electrification}. \\ https://doi.org/10.1109/TTE.2023.3313986

\bibitem[\protect\astroncite{Yang {et~al.}}{2022}]{18}
Yang L, Zhang Z, Song Y, {et~al.}, 2022.
\newblock Diffusion models: A comprehensive survey of methods and applications.
\newblock {\em arXiv preprint arXiv:220900796}. \\ https://doi.org/10.1145/3626235

\bibitem[\protect\astroncite{Yang {et~al.}}{2021}]{21}
Yang X, Wang X, Wang N, {et~al.}, 2021.
\newblock Srdn: A unified super-resolution and motion deblurring network for space image restoration.
\newblock {\em IEEE T GEOSCI REMOTE}. \\ https://doi.org/10.1109/TGRS.2021.3131264

\bibitem[\protect\astroncite{Ying {et~al.}}{2017}]{36}
Ying Z, Li G, Ren Y, {et~al.}, 2017.
\newblock A new low-light image enhancement algorithm using camera response model.
\newblock 2017 IEEE International Conference on Computer Vision Workshops (ICCVW), p.1-8. \\ https://doi.org/10.1109/ICCVW.2017.356

\bibitem[\protect\astroncite{Zhou {et~al.}}{2023}]{37}
Zhou D, Yang Z, Yang Y, 2023.
\newblock Pyramid diffusion models for low-light image enhancement.
\newblock Proceedings of the Thirty-Second International Joint Conference on Artificial Intelligence (IJCAI-23), p.1795-1803. \\ https://doi.org/10.48550/arXiv.2305.10028

\end{thebibliography}

%=========================================
%Authors can choose to use the bibliography or
%directly paste the references beginning with '\item' with the following setting.
%=========================================

\end{document}